\documentclass[]{article}
\usepackage{amsmath} 
\usepackage{amssymb} 
\usepackage{physics} 
\usepackage{mathtools} 
\usepackage{natbib}
\usepackage{graphicx} 
\usepackage{caption} 
\usepackage{booktabs} 

\usepackage{times} 
\usepackage{hyperref} 
\usepackage{enumitem} 
\usepackage{algorithm} \usepackage{algorithmic} 
\usepackage{xcolor} 

\usepackage{tensor} 
\usepackage{tikz} 
\usepackage{pgfplots} 
\pgfplotsset{compat=1.17} \usepackage{physics} 

\title{Schr\"{o}dinger AI: A Unified Spectral-Dynamical Framework for Classification, Reasoning, and Operator-Based Generalization}

\author{Truong Son Nguyen \\ Arizona State University}
\begin{document}
\maketitle
\begin{abstract}
We introduce \textbf{Schr\"{o}dinger AI}, a unified machine learning framework inspired by quantum mechanics. The system is defined by three tightly coupled components: (1) a {time-independent wave-energy solver} that treats perception and classification as spectral decomposition under a learned Hamiltonian; (2) a {time-dependent dynamical solver} governing the evolution of semantic wavefunctions over time, enabling context-aware decision revision, re-routing, and reasoning under environmental changes; and (3) a {low-rank operator calculus} that learns symbolic transformations such as modular arithmetic through learned quantum-like transition operators. Together, these components form a coherent physics-driven alternative to conventional cross-entropy training and transformer attention, providing robust generalization, interpretable semantics, and emergent topology.

Empirically, Schr\"{o}dinger AI demonstrates: (a) emergent semantic manifolds that reflect human-conceived class relations without explicit supervision; (b) dynamic reasoning that adapts to changing environments, including maze navigation with real-time potential-field perturbations; and (c) exact operator generalization on modular arithmetic tasks, where the system learns group actions and composes them across sequences far beyond training length. These results suggest a new foundational direction for machine learning, where learning is cast as discovering and navigating an underlying semantic energy landscape.
\end{abstract}

\section{Introduction}
Modern machine learning systems, especially deep neural networks trained via cross-entropy, face fundamental limitations: they struggle to represent uncertainty, adapt to new constraints at inference time, or perform symbolic reasoning without brittle memorization. The main intuition behind this limitation is that input signal are treated as discrete and output signal/ classes are treated as independent, orthogonal one-hot vector in traditional soft-max training. We propose an alternative learning paradigm grounded in quantum mechanics, where states are wavefunctions, predictions arise from energy minimization, and reasoning corresponds to dynamical evolution under a learned Hamiltonian.

We introduce \textbf{Schr\"{o}dinger AI}, a framework built around three core components:

\begin{enumerate}
\item The \textbf{Time-Independent Schr\"{o}dinger Solver}, which models classification as solving $H\psi = E\psi$ for the lowest-energy eigenstate of a learned Hamiltonian $H$. This naturally encodes semantic class topology and yields soft, physically meaningful uncertainty.

\item The \textbf{Time-Dependent Schr\"{o}dinger Dynamics}, while the time-independent version shows the most stable state that an input can shoot into, which explain the semantic ``space'' of the data, including time to the solver can boost the dynamic progression of data: since we treat input data as a wave navigating in a semantic ``space'' governed by a learned physics, the dynamic movement of the wave over time can help with: (1) correctly navigate the data modification within the ``physical space'', which help reducing the ``hallucination'' of modern generative model, and (2) slow reasoning based on input over restricted search space: if a path / energy state is blocked by constraint of the physics, the ``data wave'' automatically anneals to the updated most stable state.

\item The \textbf{Operator Calculus for Symbolic Reasoning}, which learns low-rank operators that act as quantum transitions. The system composes operators to solve tasks such as modular arithmetic, generalizing far beyond training depth.
\end{enumerate}

This unified viewpoint treats machine learning as \textit{discovering and navigating a semantic energy landscape}. We show empirically that Schr\"{o}dinger AI learns meaningful topologies, supports intuitive reasoning, and achieves precise symbolic composition.

\section{Related Work}
Our framework intersects with several research domains yet differs fundamentally from each.
\paragraph{Spectral Methods.}
Classical manifold-learning methods such as Laplacian Eigenmaps \cite{belkin2003laplacian}, Diffusion Maps \cite{coifman2006diffusion}, Spectral Clustering \cite{ng2001spectral}, and the broader diffusion geometry literature \cite{coifman2005geometric} connect data semantics to eigenfunctions of graph Laplacians. Spectral graph neural networks extend these ideas to deep learning \cite{bruna2014spectral,defferrard2017convolutional}.
Our approach differs in that the Laplacian (and more generally the Hamiltonian) is learned end-to-end through gradient descent, rather than fixed or precomputed. We also use eigenvectors not only for representation learning but as the basis for decision-making and reasoning, leveraging the ground state of a dynamically constructed Hamiltonian.

\paragraph{Neural ODEs and Continuous-Time Models.}
Neural ordinary differential equations \cite{chen2018neural} frame representation learning as continuous-time flow, and follow-up works explore controlled dynamics \cite{dupont2019augmented, rubanova2019latent}. However, these architectures typically lack explicit notions of energy, Hermitian structure, or potential fields, and therefore cannot express physically meaningful reasoning processes. Schr\"{o}dinger framework instead parameterizes a Hamiltonian operator with well-defined spectral properties, enabling wave-like propagation, tunneling, and energy-driven adaptation. 
Subsequent work has explored structured ODEs, including Hamiltonian Neural Networks (HNNs) \cite{greydanus2019hamiltonian}, which impose Hamiltonian dynamics to preserve symplectic structure in physical systems.

However, these models lack explicit potential landscapes for environment-dependent reasoning, and they do not perform eigenstate-based inference or energy-level computation. Schr\"{o}dinger framework instead constructs a Hermitian operator whose ground state encodes decisions, enabling immediate adaptation when the Hamiltonian changes, something ODE-based models do not natively support.

\paragraph{Energy-Based Models.}
Energy-based models (EBMs) \cite{lecun2006tutorial} and modern variants such as Contrastive Divergence \cite{hinton2002training}, Score-based Generative Models \cite{song2021scorebased}, and Diffusion Models \cite{ho2020denoising} represent probability via learned energies. While powerful, they do not typically learn operators governing transitions, nor do they leverage spectral structure for classification or reasoning. Closely related is the framework of Hopfield networks \cite{krotov2021large} and modern continuous Hopfield layers \cite{ramsauer2021hopfield}, which define attractor dynamics but lack Hamiltonian structure. Schr\"{o}dinger AI offers a bridge: we retain the interpretability and physical grounding of Hamiltonians while supporting both static decisions (via potentials) and dynamic replanning (via Hamiltonian updates).

\paragraph{Symbolic Reasoning and Operator Learning.}
Neural approaches to arithmetic and logic often rely on transformers \cite{vaswani2023attentionneed}, LSTMs \cite{hochreiter1997lstm}, or neural program interpreters \cite{reed2015neural}. These methods frequently struggle with systematic generalization \cite{lake2018generalization}. Some recent works attempt to learn algebraic operators \cite{banino2021pondernet,barret2018measuring}, equivalence transformations \cite{allamanis2018learning}, or group representations \cite{raghu2017expressive}, but rarely achieve perfect compositionality.
In contrast, Schr\"{o}dinger framework learns low-rank Hamiltonian operators whose action on basis states forms genuine algebraic transformations (e.g., multiplication mod $p$). This allows the system to perform multi-step symbolic reasoning via repeated operator composition, resembling group representations in physics \cite{hall2013quantum, simon2013qm} rather than sequence memorization.

\paragraph{Theoretical Foundation.} Our framework aligns with the emerging paradigm of Operator-Based Machine Intelligence, which establishes a formal Hilbert space framework for spectral learning and symbolic reasoning~\cite{kiruluta2025operatorbased}. While ~\cite{kiruluta2025operatorbased} provide a rigorous mathematical treatment of Hilbertian signal processing and the algebraic properties of operators, Schrödinger AI extends this foundational vision by grounding it in the specific dynamics of non-relativistic quantum mechanics.

Specifically, we move from a general operator framework to a physics-driven implementation defined by a learned Hamiltonian $(H=m^{-1}K+V)$. Unlike purely algebraic operator models, Schrödinger AI utilizes a mass term ($m$) to govern the balance between quantum tunneling and classical stability and leverages Hamiltonian updates to achieve real-time adiabatic re-planning in changing environments. Thus, while the former provides the foundational algebra for Hilbert-space AI, our work provides the dynamical engine for energy-driven reasoning and perception.

\section{The Schr\"{o}dinger AI Framework}
We now formalize our three components.

\subsection{Time-Independent Schr\"{o}dinger Solver for Classification}
Given an input representation $x$, we define a learned Hamiltonian $H(x)$ acting on a finite-dimensional Hilbert space. The model predicts the class by solving the eigenvalue problem:

\begin{equation}
H(x)\psi = E\psi,
\end{equation}

where $E$ and $\psi$ denote energy levels and eigenstates. The lowest-energy eigenstate corresponds to the most probable class. Unlike softmax logits, the energy spectrum encodes semantic proximity: classes closer in topology correspond to similar eigenstates, allowing natural uncertainty representation.

We introduce a \textbf{mass term} $m$ governing stability: heavy signals (clear inputs) collapse into deep minima, while light/noisy signals remain diffuse and may anneal between nearby states.

In our practical implementation of Schrödinger AI for CIFAR-10, the time-independent solver is instantiated as a Hamiltonian head on top of a Vision Transformer (ViT) backbone. The ViT produces a CLS embedding that parameterizes all components of the Hamiltonian governing the semantic energy landscape. Classification is performed by computing the ground-state eigenvector of this Hamiltonian, and interpreting its squared magnitude as a probability distribution over classes.

\subsubsection{From Image to Quantum State: The ViT Encoder
}

An input image is tokenized into non-overlapping patches and passed through a standard Transformer encoder. The final CLS embedding $h\in \mathbb{R}^d$ summarizes the perceptual content of the image.

From this representation, we compute three quantities that define the Hamiltonian:

\textit{Class potential vector} \[V=W_V\cdot h\]

where each entry $V_i$ represents the intrinsic potential energy associated with class i.

\textit{Mass} The CLS embedding is mapped to a scalar mass:
\[m=0.05+9.95\cdot\sigma(W\cdot M\cdot h)\]

which determines the strength of the kinetic term and therefore governs how “quantum” the system behaves.

This mass term corresponds to kinetic energy contribution, where a small mass directly shows a broad, annealable wave functions that can tunnel to different energy states, and large mass shows a narrow band wave. This broad and narrow band waves translate to uncertainty in prediction of certain class, where classes translate exactly to energy levels.

\textit{Global class topology matrix.} 
A learnable matrix $\Theta \in \mathbb{R}^{C\times C}$
 stores latent semantic affinities between classes. To ensure physically meaningful structure, we symmetrize it to $W=\frac{1}{2}(\Theta^T+\Theta)$. We also enforce non-negativity of energy via softplus, remove self-connections, and construct the graph Laplacian:

\[K=D - W\],

where $D$ is the degree matrix. The diagonal $D$ can be interpreted as the class ``inertia'' making the prediction stable, while the learned $W$ identify the tunnel possibility, give the model aware of possibilities of different prediction for the input. This learned Laplacian $K$ acts as a kinetic energy operator, defining which classes are ``adjacent'' or ``easy to tunnel between.''

\subsubsection{Constructing the Hamiltonian}

The Hamiltonian for each sample in the batch is:

\[H=m^{-1}(K+\epsilon I)+\text{diag}(V)\],
where $\epsilon$ is a small stabilization constant. We use the potential term diag$(V)$ to attract the wavefunction toward class-specific likelihood basins, and the kinetic term $m^{-1}K$ spreads probability mass along the learned topology, promoting uncertainty between semantically related classes (cat–dog, car–truck, plane–bird, etc.). The mass controls the balance between exploration (quantum behavior: tunneling) and collapse (classical behavior: staying until enough kinetic energy).

\begin{figure}
    \centering
    \includegraphics[width=0.8\linewidth]{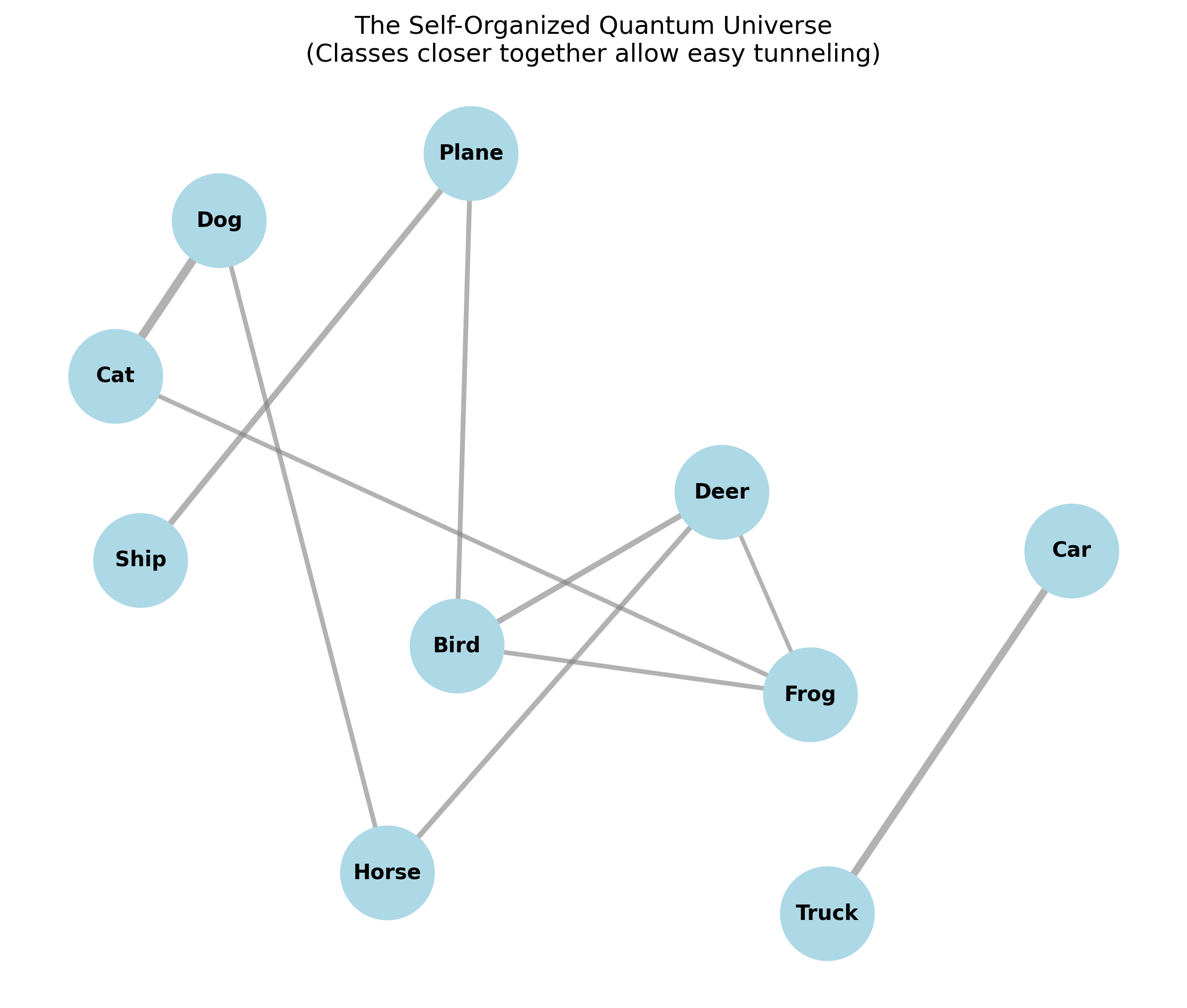}
    \caption{Learned inherent relationship among classes of a learned Schr\"{o}dinger classifier}
    \label{fig:relationship}
\end{figure}

Unlike softmax classifiers, which treats classes as isolated islands, our model learns the connectivity between classes and allow input to ``leak'' to a different class if its mass is small. This results in the interesting implicitly learned structure of connectivity between class that are semantically close in the data. We visualize the learned Laplacian $K$ to show the connectivity and show it in Figure ~\ref{fig:relationship}. It shows that in even in a small CIFAR10 dataset of 50000 images, the model can learn the connection between vehicle (car-truck), pets (dog-cat strong connections), animals (cat-dog-horse-deer-frog cycle), open space background vehicle (ship/plane) and flying things (plane-bird).

\subsubsection{Solving for the Ground State}

For each sample at output, we compute the full eigen-decomposition: $H\psi = E\psi$ and find the lowest-energy eigenvector $\psi_0$, and defines the perceptual state of the input: $p_i=|\psi_{0,i}|^2$.
There are some interesting properties of this distribution:
\begin{itemize}
\item Normalized automatically by quantum mechanics,
\item Smooth across semantically related classes,
\item Reflective of uncertainty, because topology and mass shape the wavefunction spread.
\end{itemize}

For classification use, we compute $\hat{y} = \text{argmax}(p_i)$, while the topological structure learned in 
$K$ naturally leads to interpretable uncertainty patterns, e.g., a blurry cat image may place mass across cat, dog, horse, depending on learned adjacency.

\subsubsection{Regularization and Stability}

To avoid degenerate solutions, we apply the following regularization:
\begin{itemize}
\item A mass penalty to keep mass small, preserving quantum behavior.
\item A sparsity penalty on $K$ to encourage meaningful, nontrivial class topology.
\end{itemize}

These constraints ensure the Hamiltonian expresses a semantic energy landscape rather than collapsing to a trivial diagonal classifier.

\subsection{Time-Dependent Dynamics for Rerouting}

When the environment of the particle changed, we directly compute the lowest-energy eigenstate of the updated Hamiltonian \[H'(t) = \frac{1}{M}K+\text{diag}(V(t))\]
where the potential term $V(t)$ changes in response to the newly observations. The ground state $\psi_0(t)$ represents the system's current belief over feasible future states. When the environment changes, the Hamiltonian changes immediately, and the recomputed ground state produces a new decision distribution without requiring any retraining or gradient updates.

This mechanism is useful in applications: the update happens at test time according to some update in environment (i.e. data), not update in model weights. This helps the model to update itself to new rules, regulations, or observations at test time. We show this phenomenon particularly in two experiments: rule updating and dynamic rerouting based on environment.


\subsection{Auto Updating as Stable Energy-Seeker}
\label{sec:reasoning-quantum}

We now describe, in operational and quantum-mechanical terms, how the Schr\"{o}dinger AI performs rerouting in the rule enforcing and navigation tasks presented in this work. The treatment below emphasizes (i) the precise mathematical objects we learn, (ii) how decisions are extracted from quantum-like inference, (iii) how online adaptation (``reasoning'') is achieved by Hamiltonian updates, and (iv) the role of inductive priors and annealing schedules that proved critical in practice.

\paragraph{State space and wavefunction.}
Let the environment be represented by a discrete set of basis states $\{1,…,S\}$. We associate a complex amplitude 
$\psi(s)\in \mathbb{C}$ to each basis state and collect these into a state vector $\psi \in \mathbb{C}^S$. The model interprets the squared amplitude $|\psi(s)|^2$ as the agent's current belief or utility mass over candidate states, directly analogous to the Born rule in non-relativistic quantum mechanics.

\paragraph{Learned Hamiltonian and its components.}
Reasoning is governed by a learned, sample-dependent Hamiltonian of the form
\begin{equation}
\label{eq:H-def}
H = \frac{1}{m}K + \operatorname{diag}(V),
\end{equation}
where
\begin{itemize}
\item $K\in \mathbb{R}^{S\times S}$ is a symmetric graph Laplacian (kinetic operator) encoding connectivity and tunneling amplitudes between states; in implementation $K=D-W$ with a nonnegative symmetric adjacency $W$.
\item $V\in \mathbb{R}^S$ is a learned potential vector (diagonal operator) produced conditionally on the current observation and goal.
\item $m>0$ is a learned or inferred scalar mass that controls the relative strength of kinetic spreading (tunneling) versus localization by the potential.
\end{itemize}
Both $W$ (hence $K$) and $V$ are produced by neural modules conditioned on perceptual/contextual embeddings; $K$ is constrained to be symmetric so that $H$ is Hermitian and its spectrum is real.

\paragraph{Decision extraction via ground-state inference.}
At inference time, the model computes the lowest-energy eigenpair of $H$:
\begin{equation}
H\psi_{0}=E_{0}\psi_{0},\qquad E_{0}=\min_{|\psi|=1}\psi^{\dagger}H\psi.
\end{equation}
We use the ground-state vector $\psi_0$ as the model's slow (deliberative) belief; the decision distribution is taken proportional to $|\psi_0(s)|^2$. This single-step eigen-solve replaces multi-step rollout or value-iteration commonly used in planning: the ground state is the variational minimizer of expected energy and therefore concentrates on low-energy (high-utility, reachable) regions of state space.


\paragraph{Online adaptation as Hamiltonian update (time-dependent behavior).}
Rather than integrating the time-dependent Schr\"{o}dinger equation, the implemented reasoning procedure realizes time-varying behavior by \emph{Hamiltonian adaptation and repeated ground-state inference}. Concretely, when new sensory evidence arrives (a blocked cell, a moving obstacle, or a changed goal), we update the potential \[V\mapsto V'=V+\Delta V\] rebuild $H'=(1/m)K + \text{diag}(V)'$, and recompute the ground state $\psi_0'$. The resulting redistribution of $|\psi_0'|^2$ corresponds to instantaneous re-planning: probability mass vacates newly penalized states and flows into alternative basins allowed by 
$K$. This mechanism yields the empirical property that \emph{the agent reconfigures its decisions without any gradient-based retraining}, producing rapid rerouting in the maze experiments.

\paragraph{Local neighbor selection from global wavefunction.}
Because $\psi_0$ is a global object, we must convert a preferred global target $s'=\texttt{argmax}_s|\psi_0(s)|^2$ into an immediate local action (one of the agent's legal neighbors). We implemented a reachability-aware mapping: for each legal neighbor $n$ of the current cell we compute a local score \[\texttt{score}(n)=\sum\limits_{s\in\mathcal{R}(n)}|\psi_0(s)|^2\], where $\mathcal{R}(n)$ is a bounded-depth reachable set (BFS radius $d$) from $n$. The neighbor maximizing this score is selected. This procedure explicitly favors neighbors that lead into high-probability basins under $\psi_0$ while avoiding blocked neighbors and unreachable targets; it is a principled discrete analogue of following the local probability flux suggested by the wavefunction.

\paragraph{Topology prior and annealing ($\alpha$ schedule).}
To stabilize learning and preserve environmental geometry, the learned adjacency matrix $W_\text{learned}$ is combined with a fixed grid adjacency prior $W_\text{grid}$ via a convex mixing coefficient $\alpha$:

\[W=\alpha\cdot W_\text{grid}+(1-\alpha)\cdot W_\text{learned}, K=D-W. \]

During training we anneal $\alpha$ from a physics-dominated regime ($\alpha=0.9$) to a data-dominated regime ($\alpha=0$). This curriculum enforces correct local connectivity early (forcing model to follow the correct graph traversal rules) while permitting semantic connectivity to emerge later: once the model learned the rule to traverse in the graph, they are free to choose to flow in which possible directions. 

\subsection{Surgical Rule Injection via Hamiltonian Perturbations}

A unique advantage of the Schr\"{o}dinger AI framework is the ability to perform precise, zero-shot model editing without gradient-based fine-tuning. We introduce Surgical Rule Injection, a method that modifies the learned Hamiltonian $H$ to re-route specific semantic classifications while preserving global integrity.

To map a source class $c_{from}$ to a target class $c_{to}$, we perturb the Hamiltonian $H$ for an input $x$ as follows:
\begin{itemize}
    \item Source Repulsion: We increase the potential $V[c_{from}]$ by a factor $\alpha$, creating an energetic barrier that discourages the ground state from occupying the source well.

    \item Transition Tunneling: we introduce off-diagonal "tunneling" terms $H_{c_{from},c_{to}}$ and $H_{c_{to},c_{from}}$. These terms act as transition amplitudes, physically ``connecting'' the two semantic wells in the Hilbert space.

    \item Potential Anchoring: To prevent semantic leakage, we apply an Anchoring Potential to all classes $c\notin\{c_{from},c_{to}\}$. By deepening the potential wells of the non-target classes, we ensure that the global stability of the model remains intact, preventing the "surgical" change from degrading the accuracy of unrelated categories.
\end{itemize}

This process is governed by a sigmoid-based gating mechanism G(x) that determines the ``strength'' of the rule application based on the model’s initial confidence, ensuring the rule is only applied to relevant inputs.

\subsection{Operator Calculus for Symbolic Reasoning}

A distinguishing feature of Schr\"{o}dinger AI is its ability to learn and manipulate symbolic transformations using the same physical principles that govern its perceptual and dynamical behavior. Instead of learning arithmetic by memorization or pattern matching, the system learns a bank of quantum-like operators that encode algebraic structure directly through wave evolution.

In the modular arithmetic experiments (e.g., multiplication modulo 13), the model learns a family of low-rank matrices $O_b \in \mathbb{R}^{C\times C}$, one for each symbolic input $b$. These operators act on a probability wave over symbolic states. A state $\psi$ evolves over time under operator $O_b$ as: \[\psi' = \psi O_b \], This evolution rule is analogous to the application of a quantum transition operator, where the wavefunction is transported across discrete states according to the learned algebraic relations.

\subsubsection{Training procedure.} In our toy example, we train the model to learn ``modulo multiplication'': each arithmetic expression $a\times b = c \mod 13$ is treated as a transition: \[O_be_a = e_c\],
where: $e_a, e_c$ are one-hot basis of input $a$ and $c$ respectively and b is the multiplier (or general operator symbol),

Instead of predicting a label, the model must learn an operator whose action transports the wavefunction from $\psi_a$ to $\psi_c$.

To ensure that these operators generalize compositionally, we introduce two architectural constraints:

\begin{enumerate}
\item \textbf{Operator depends on the second operand only}: the operator $O_b$ must be a function of b alone, and not of the left operand a. This enforces a strong inductive bias: the operator must implement a true algebraic action, not memorize pairwise mappings.

\item \textbf{Low-rank factorization}. Each operator is parameterized as: $O_b = L\cdot$diag$(z_b) \cdot R^T$, 
where $L$ and $R$ are global matrices and $z_b$ is a learned embedding for symbol $b$.
\end{enumerate}

This dramatically reduces parameter count while preserving expressivity.

\begin{figure}
    \centering
    \includegraphics[width=0.9\linewidth]{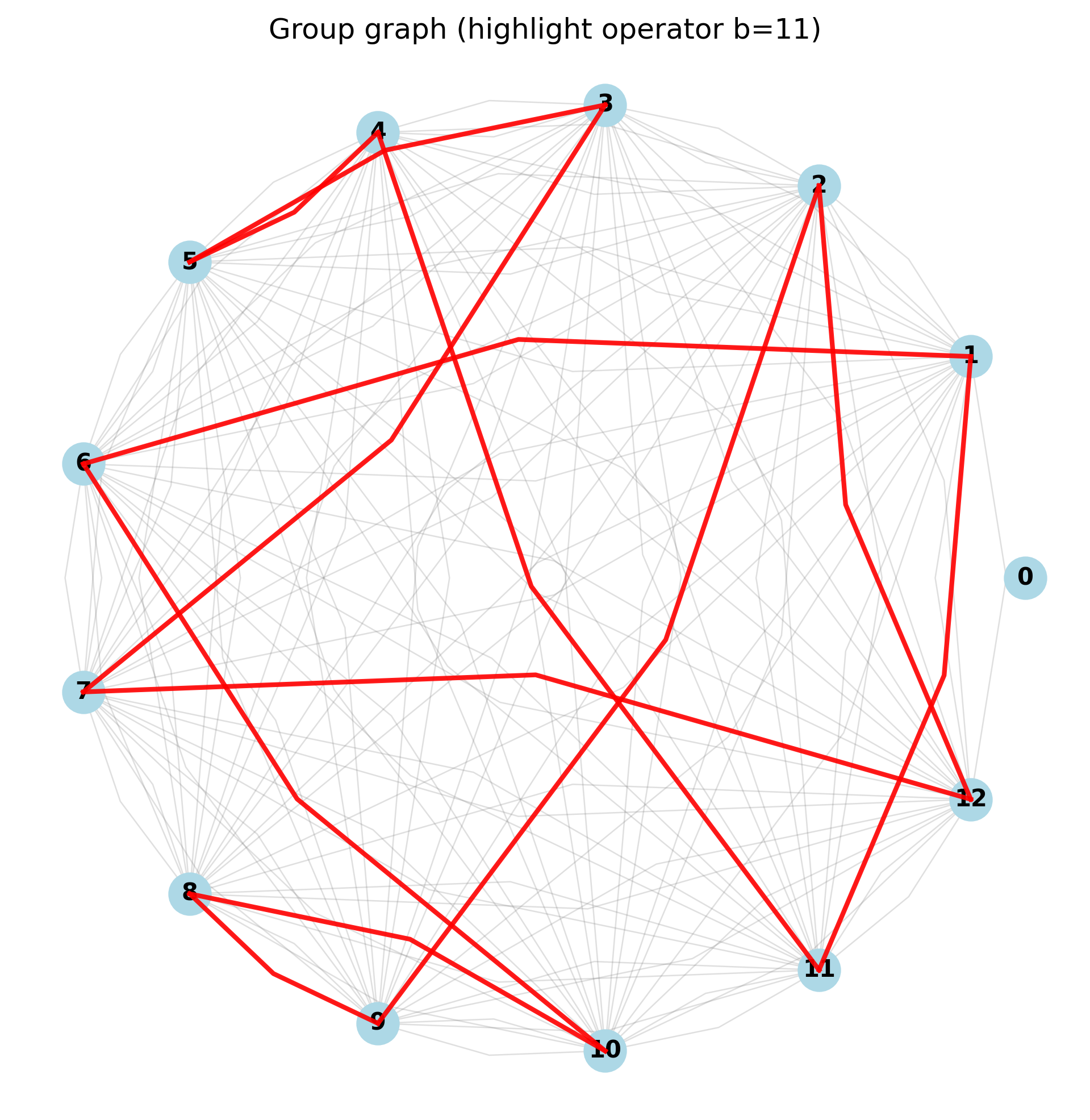}
    \caption{Visualization of the group connection learned by Schr\"{o}dinger reasoning model on $O_{11}$. It shows node $0$ is isolated while node $i$ and $j$ are connected if either $11*i\equiv j \mod 13 $ or $11*j\equiv i \mod 13$}
    \label{fig:learned_group}
\end{figure}

This factorization encourages the operators to share structure and align with the underlying group representation. We visualize the the group connections among elements in group of modulo $13$ and operator $O_{11}$. The resulted visualization is shown in Figure ~\ref{fig:learned_group}.

\subsubsection{Compositional Reasoning Through Operator Chains}

Once operators are learned, the system can perform multi-step reasoning, i.e. computing $a_1\times \dots\times a_n = (((e_{a_1}O_{a_2})\dots) O_{a_n})$

This corresponds to applying the multiplication rule repeatedly. A key emergent behavior is that the model learns operators satisfying: \[O_a\cdot O_b = O_{ab\mod p}\]

This is the characteristic property of a group representation.

Empirically, the learned operators:

\begin{itemize}
\item generalize perfectly to long chains unseen during training,
\item behave as nearly permutation matrices on the symbolic states,
\item obey the homomorphism property to high numerical precision.
\end{itemize}
While the typical neural networks fail to learn such algebraic structure reliably, even with extensive training, Schr\"{o}dinger scheme does it reliably with just a limited number of training.

\subsection{Schr\"{o}dinger Agent: Combining Operator Calculus and Dynamic Reasoning}
Once we have operator calculus learner and environment energy learner, it is sensible to combine them to form an agent: An AI system that is aware of both ``what I can do'' (operator calculus) and ``where I am/ where I should go'' (environment energy) and can integrate both perceptions to make optimal decision. We come up with a 3-stage approach for training Schr\"{o}dinger AI Agent:
\begin{enumerate}
    \item Stage 1: Action Awareness: We explicitly train the awareness of actions to the agent without any explicit rule of environment. In our maze solving example, this is where the agent learns what it can do (up/down/left/right) movement without any obstacle introduced.
    \item Stage 2: Environment Awareness (Perception): We train the model to perceive its environment by introducing multiple different environment and train its Hamiltonian to calculate the energy landscape of those with respect to different positions.
    \item Stage 3: Action-Perception alignment: Now we have agent understand both actions and directions, we will align them together that agent chooses an action, the action feat will be fit with both the actual outcome (so that the agent does not forget what the action it can do) and the ``stableness'' of the next position according to the environment perception.
\end{enumerate}

\subsubsection{Dual-system Reasoning Designed with Schr\"{o}dinger Agent}
\label{sec:reasoning-dual}

In addition to training the agent that can both act and perceive, we would like to create a ``fast track'' that bypass reasoning about environment so that it reflect the habitual thinking of our cognition. With Schr\"{o}dinger AI, this is simply just appending a new, additional, light habitual head to the model and train it during the stage 3 phase along with the action-perception alignment. The habitual head simply choose one of the four direction using normal soft-max. We directly use the environment perception to provide the target for this habitual head. Then, the task of the habitual head is just outputting the most probable lowest stage. Since it is not aware of the actions it can take, it is different (and faster) than the operator calculus perceiver in the agent

\section{Experiment Result}
We evaluate Schr\"{o}dinger AI on multiple categories of tasks as described in the previous section. The implementation can be found at \url{https://github.com/sonnguyenasu/SchrodingerAI}

\subsection{Semantic Classification via Ground-State Discovery}
\begin{figure}
    \centering
    \includegraphics[width=\linewidth]{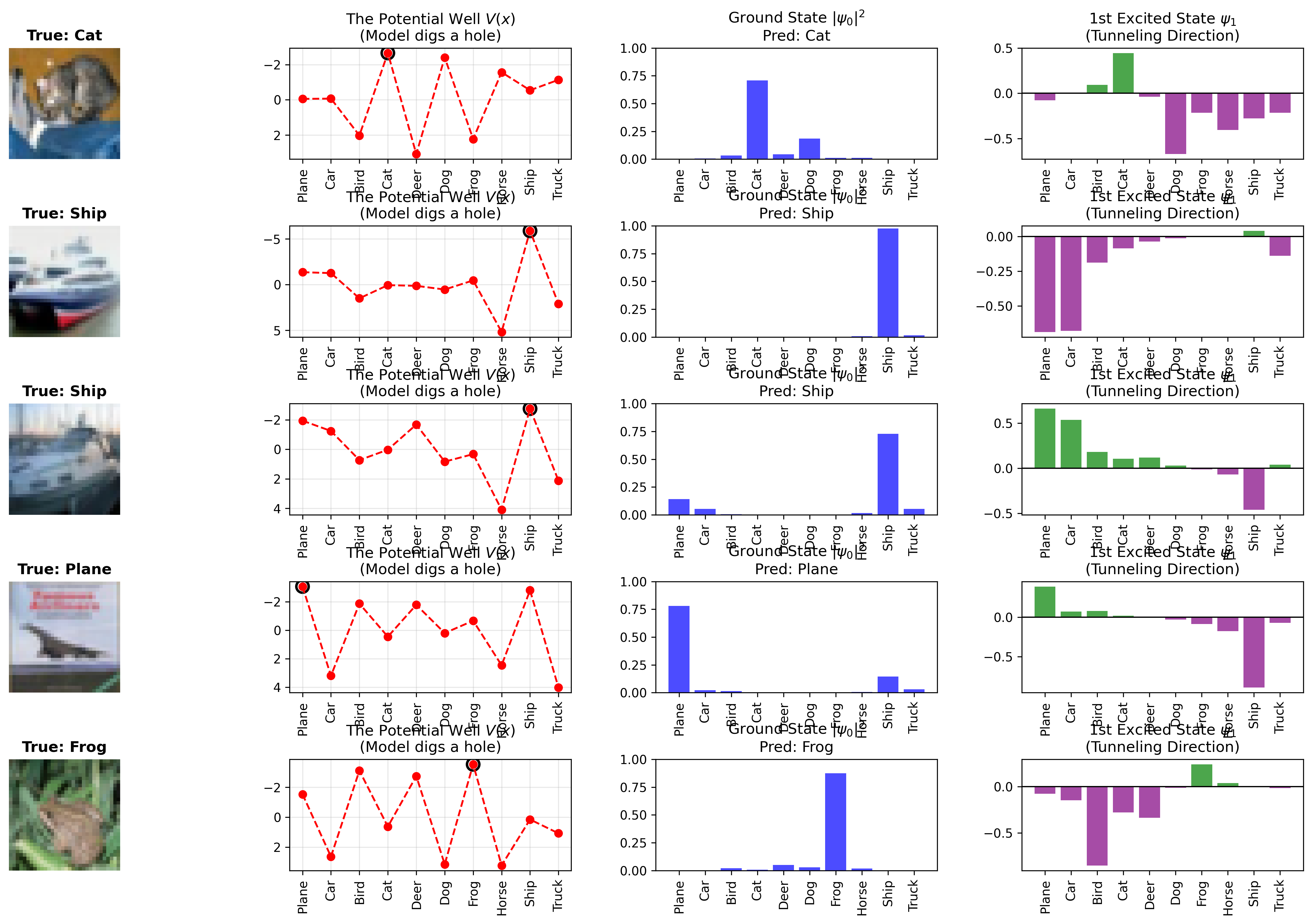}
    \caption{Sampled output of the trained Schr\"{o}dinger classifier. First column: Sampled input, Second column: Potential Well $V(x)$, Third column: Prediction $|\psi_{i,0}|^2$, Forth column: Tunneling direction (next possible guess).}
    \label{fig:test_classifier}
\end{figure}

Using a lightweight ViT model with a spectral head, we achieved 76\% accuracy on CIFAR-10 after 40 epochs, with train and test performance matched. The confusion graph reveals emergent topology: e.g., Cat-Dog proximity, Car-Truck grouping, Bird-Plane affinity. These arise without explicit structural priors. The sampled classified results are shown in Figure ~\ref{fig:test_classifier}.

\begin{figure}
    \centering
    \includegraphics[width=\linewidth]{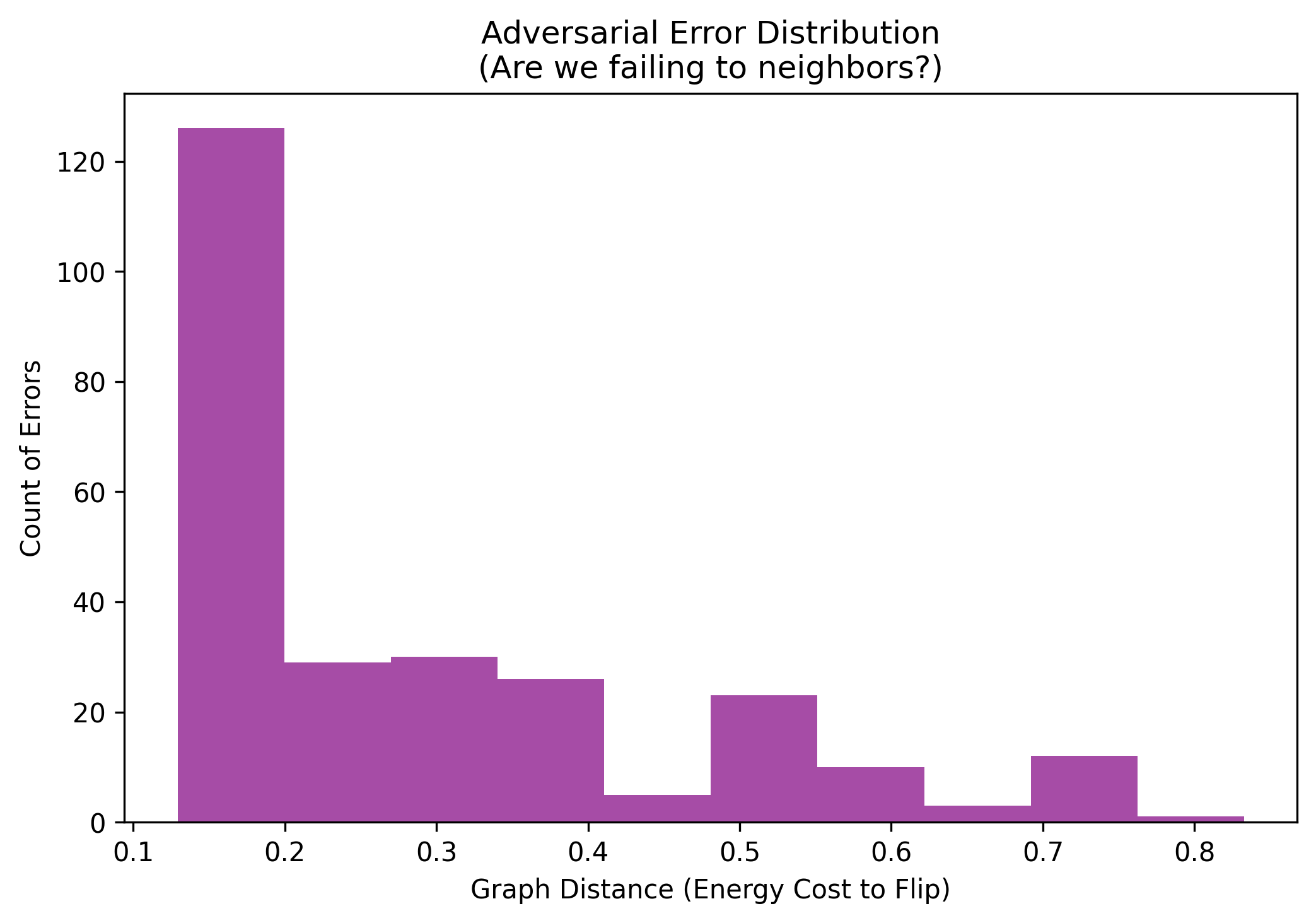}
    \caption{Distribution of the distance (traversed graph edges) from original label to the failed output target.}
    \label{fig:pgd}
\end{figure}

To further shows the robustness of the learned model, we implemented a projected gradient descent (PGD) attack~\cite{pgdattack} on the learned model. The result is that the learned model accuracy dropped from 76.41\% to 17.50\%, which retains roughly $23\%$ of its original capability. However, the changed label of the input image after the attack are interpretable: The image is highly likely change to its neighbor in the semantic graph (or the one with tunneling possible) rather than an irrelevant output. This phenomenon of the model under attack indicates a true understanding of the semantic of the input data. We show the distribution of the failed input after PGD attack in Figure ~\ref{fig:pgd}.

\subsection{Zero-Shot Rule Compliance (Classes Rerouting)}

To evaluate the flexibility of our physics-driven framework, we conducted a "Surgical Rerouting" experiment. We tasked the model with instantly re-mapping all image of one label (Class $i$, e.g. ``dog'') predictions to image of another label (Class $j$, e.g. ``cat'') without updating any neural network weights.

Using the Anchored Surgical Tunnel method, we injected transition amplitudes into the Hamiltonian at inference time. We measured two key metrics: Rule Compliance (percentage of Dogs successfully identified as Cats) and Global Stability (accuracy retention for the other 9 classes). We show the confusion matrix of Rule Compliance in Table ~\ref{tab:reroute1} and Global Stability in Table ~\ref{tab:reroute2}

\begin{table}[h]
\centering
\small
\begin{tabular}{l|cccccccccc}
\hline
\textbf{From \textbackslash To} & \textbf{Plane} & \textbf{Car} & \textbf{Bird} & \textbf{Cat} & \textbf{Deer} & \textbf{Dog} & \textbf{Frog} & \textbf{Horse} & \textbf{Ship} & \textbf{Truck} \\
\hline
\textbf{Plane} & - & 85.9 & 87.3 & 86.8 & 86.4 & 85.7 & 85.5 & 86.1 & 86.2 & 87.7 \\
\textbf{Car} & 88.5 & - & 87.2 & 87.6 & 87.1 & 87.4 & 86.8 & 87.9 & 87.6 & 95.5 \\
\textbf{Bird} & 76.0 & 73.3 & - & 78.1 & 79.9 & 78.7 & 75.2 & 77.3 & 73.4 & 74.4 \\
\textbf{Cat} & 78.4 & 78.4 & 80.4 & - & 80.6 & 84.5 & 79.7 & 81.0 & 78.2 & 80.2 \\
\textbf{Deer} & 80.0 & 79.3 & 82.5 & 82.6 & - & 81.5 & 81.8 & 84.6 & 79.4 & 79.5 \\
\textbf{Dog} & 84.8 & 84.4 & 85.8 & 88.5 & 87.3 & - & 85.0 & 88.8 & 84.3 & 85.3 \\
\textbf{Frog} & 83.1 & 83.1 & 86.2 & 87.8 & 85.9 & 85.4 & - & 83.6 & 82.8 & 83.7 \\
\textbf{Horse} & 88.6 & 87.4 & 88.8 & 90.0 & 90.0 & 90.2 & 87.4 & - & 87.6 & 89.3 \\
\textbf{Ship} & 88.2 & 85.0 & 82.6 & 83.1 & 82.7 & 82.3 & 82.0 & 82.3 & - & 86.4 \\
\textbf{Truck} & 94.0 & 94.8 & 92.6 & 93.1 & 92.6 & 92.6 & 92.3 & 92.8 & 92.8 & - \\
\hline
\end{tabular}
\caption{Rule Compliance Matrix for Zero-Shot Surgical Rerouting. Each cell represents the percentage of successful transitions from the source class (row) to the target class (column) using Hamiltonian tunneling perturbations.}
\label{tab:reroute1}
\end{table}

\begin{table}[h]
\centering
\small
\begin{tabular}{l|cccccccccc}
\hline
\textbf{From \textbackslash To} & \textbf{Plane} & \textbf{Car} & \textbf{Bird} & \textbf{Cat} & \textbf{Deer} & \textbf{Dog} & \textbf{Frog} & \textbf{Horse} & \textbf{Ship} & \textbf{Truck} \\
\hline
\textbf{Plane} & - & 72.7 & 73.1 & 72.4 & 72.5 & 72.4 & 72.4 & 72.5 & 74.1 & 72.7 \\
\textbf{Car} & 72.8 & - & 72.6 & 72.6 & 72.6 & 72.6 & 72.6 & 72.6 & 73.2 & 73.6 \\
\textbf{Bird} & 74.4 & 73.6 & - & 74.4 & 74.9 & 74.0 & 74.7 & 73.9 & 73.8 & 73.6 \\
\textbf{Cat} & 72.7 & 72.3 & 73.6 & - & 73.0 & 76.3 & 73.8 & 72.8 & 72.4 & 72.3 \\
\textbf{Deer} & 73.5 & 73.2 & 75.0 & 73.8 & - & 73.8 & 74.0 & 73.9 & 73.3 & 73.2 \\
\textbf{Dog} & 71.6 & 71.4 & 72.9 & 76.3 & 72.0 & - & 72.3 & 72.3 & 71.5 & 71.5 \\
\textbf{Frog} & 73.5 & 73.4 & 73.9 & 74.0 & 73.9 & 73.5 & - & 73.3 & 73.4 & 73.4 \\
\textbf{Horse} & 72.0 & 71.9 & 72.5 & 72.4 & 73.5 & 72.9 & 71.9 & - & 71.9 & 71.9 \\
\textbf{Ship} & 73.9 & 73.4 & 73.3 & 73.4 & 73.3 & 73.3 & 73.3 & 73.3 & - & 73.4 \\
\textbf{Truck} & 71.6 & 73.2 & 71.1 & 71.3 & 71.0 & 71.1 & 71.1 & 71.2 & 71.9 & - \\
\hline
\end{tabular}
\caption{Global Stability Matrix. Each cell represents the accuracy of all classes \textit{except} the source class when the corresponding row-to-column surgical rule is active. The baseline accuracy (no rules) is approximately 76.5\%, indicating that surgical tunneling imposes a consistent but manageable 'stability tax' of $\sim$2-3\% on the overall manifold.}
\label{tab:reroute2}
\end{table}


\subsection{Dynamic Reasoning in Changing Environments}

We evaluate the adaptive reasoning capabilities of Schr\"odinger AI in a $10\times10$ grid navigation task. Unlike traditional reinforcement learning policies that rely on fixed pattern matching, our framework utilizes a dual-system architecture that couples a fast discriminative head (System 1) with a deliberative ground-state solver (System 2).

\subsubsection{Empirical Results: Efficient Navigation at New Environment}
We tested the model on 80 freshly generated test mazes containing real-time potential-field perturbations. The Energy Stable seeker model achieved a $97.5\%$ success rate, significantly outperforming the softmax-based predictor baseline, which dropped to $60\%$ accuracy due to overfitting on training maps. 

    

\begin{figure}
    \centering
    \includegraphics[width=\linewidth]{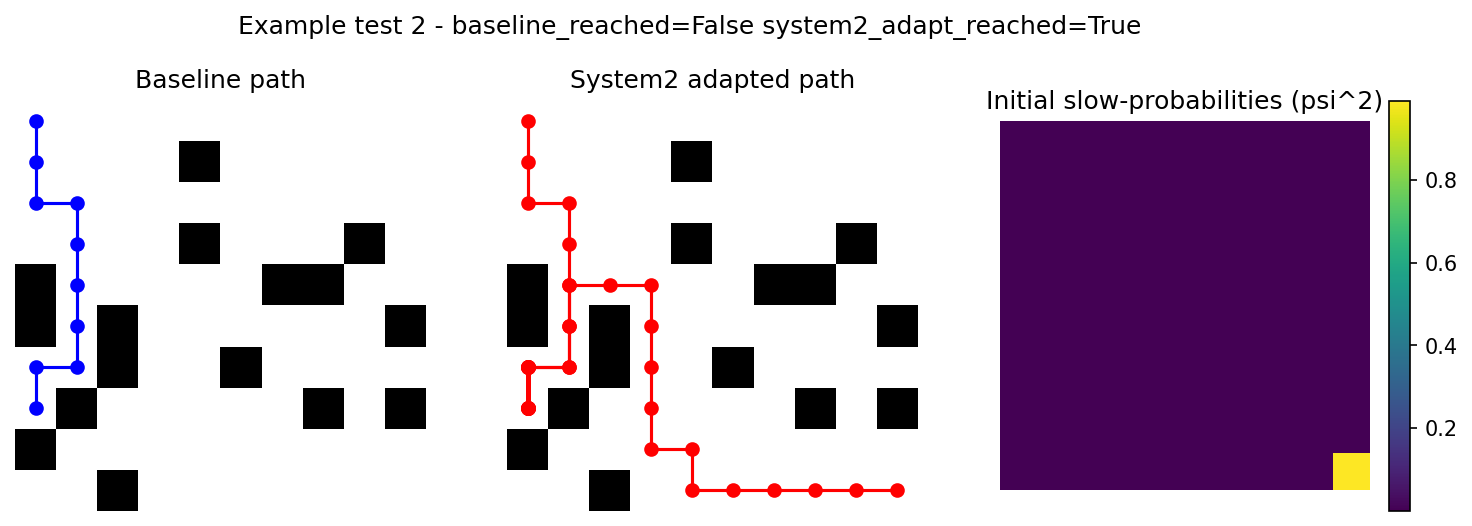}
    \caption{Visualization example of the planning of Our Schr\"{o}dinger AI Energy Seeker outperforms the classical LSTM route planner (fast) and energy heat map showing slow model awareness of the final goal, and that helps the agent naturally flows to the goal, unlike habitual model.}
    \label{fig:replanning}
\end{figure}

As illustrated in Figure \ref{fig:replanning}, Energy seeker model maintains a clear global awareness of the final goal through its lowest-energy eigenstate, enabling it to bypass local traps where System 1 often fails. This confirms that the Hamiltonian acts as a robust semantic landscape that provides corrective feedback to the fast-reasoning pathway.

\subsection{Symbolic Reasoning via Operator Composition}
We train low-rank operators to model multiplication modulo 13. The model achieves perfect pairwise accuracy and composes operators over sequences up to length 20 with zero error. The learned operators nearly satisfy the group homomorphism property, enabling generalizable symbolic reasoning. In addition, the model learn the underlying group structure with respect to multiplication where the visualized connection only highlightened at pair $(i,j)$ where $e_j=e_iO_b$ or vice-versa as shown in Figure ~\ref{fig:learned_group}.

In Table ~\ref{tab:per_op_acc} we show a pair-wise held-out accuracy per operator $(O_{b\in[1,12]})$ .We also test the performance of the model on variable chain length of multiplication modulo 13. The results are shown in Table ~\ref{tab:chain}
\begin{table}
\caption{Pairwise held-out accuracy per operator.}
\label{tab:per_op_acc}
\centering
\begin{tabular}{rrrrrrrrrrrrr}
\toprule
operator & 1 & 2 & 3 & 4 & 5 & 6 & 7 & 8 & 9 & 10 & 11 & 12 \\
\midrule
accuracy & 1.0 & 1.0 &1.0 &1.0 &1.0 &1.0 &1.0 &1.0 &1.0 &1.0 &1.0 & 1.0 \\
\bottomrule
\end{tabular}
\end{table}

\begin{table}
\caption{Composition generalization (accuracy vs chain length).}
\label{tab:chain}
\centering
\begin{tabular}{rrrrrrrrrrr}
\toprule
chain length & 2 & 4 & 6 & 8 & 10 & 12 & 14 & 16 & 18 & 20 \\
\midrule
accuracy & 1.0 & 1.0 & 1.0 &1.0 &1.0 &1.0 &1.0 &1.0 &1.0 &1.0\\
\bottomrule
\end{tabular}
\end{table}

\subsection{Maze solver using dual thinking system}
As discussed in Section ~\ref{sec:reasoning-dual}, we can combine Operator Calculus learning agent with Environment learning agent and a simple policy predicting agent to end up with a dual system of thinking. In this experiment, we use the same maze solving task as in environment learning, but now we also train the operator calculus along with it. The result is surprising: While the environment stable seeker in the previous experiment achieve 97.5\% accuracy on finding the path it hasn't seen, the dual system with reasoning and acting awareness jumps to 99.5\% path finding capacity. This is because the agent not only look at the next visible way to flow, but also have ability to ``see'' available paths using the action awareness. We visualize an example of this dual-system path finding in Figure ~\ref{fig:dual-system} and multiple other path finding in Appendix. A surprising impression is that the model can find it way to the goal even when it get lost and has to replan for a lot of times, making the decision path look almost like an exhaustive depth-first-search.

\begin{figure}
    \centering
    \includegraphics[width=0.95\linewidth]{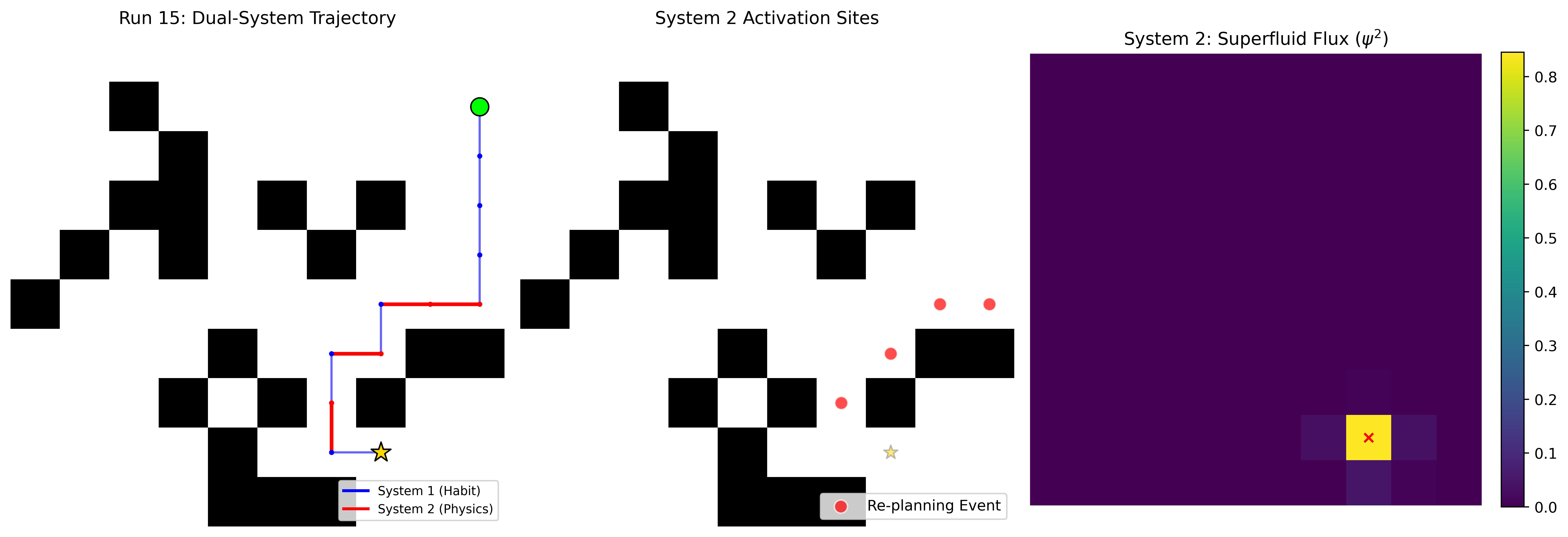}
    \includegraphics[width=0.95\linewidth]{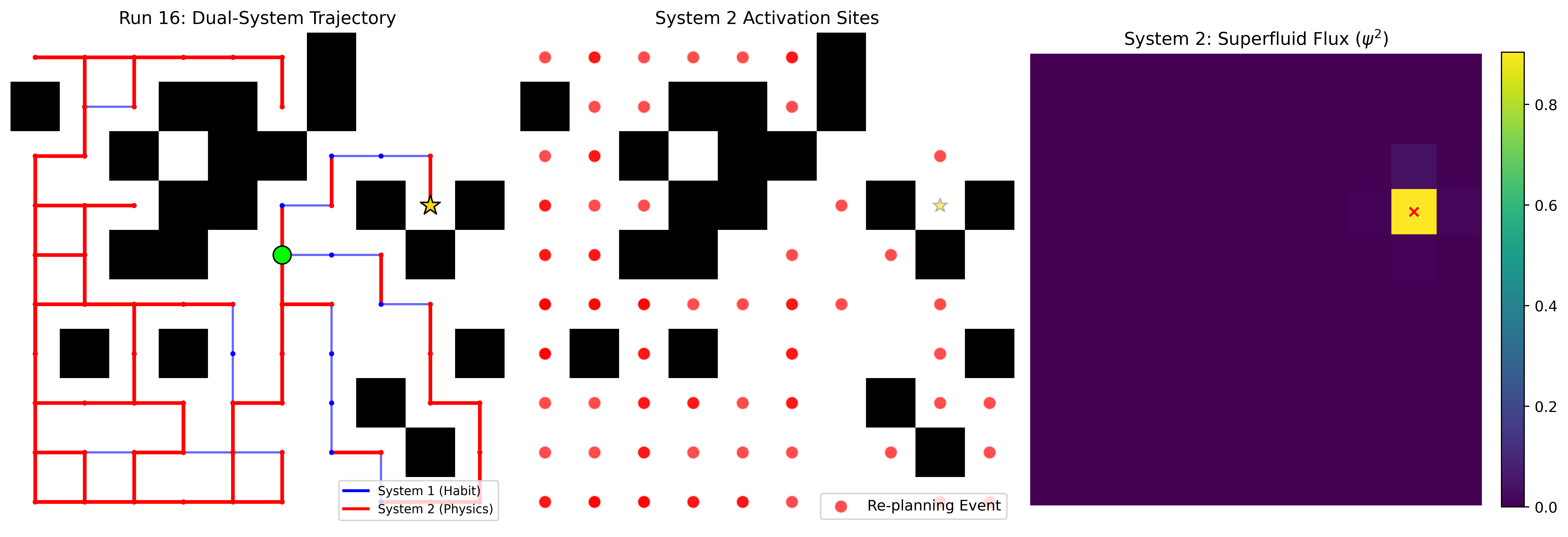}
    \caption{Dual-system path finding sample. Up: ``easy'' maze where the agent can find answer directly. Down: ``hard'' maze where the agent gets lost but eventually finds the way out.}
    \label{fig:dual-system}
\end{figure}

\section{Discussion}
Our results indicate that spectral and dynamical formulations can unify classification, reasoning, and symbolic computation within a single mathematical framework. The Hamiltonian acts as a learned semantic landscape; wavefunctions represent belief states; operator actions encode transformations; and annealing provides robustness and adaptivity.


\section{Conclusion}
We present Schr\"{o}dinger AI, a unified spectral-dynamical operator framework capable of classification, reasoning, and symbolic generalization. By grounding machine learning in physical principles of energy minimization, wave dynamics, and operator algebra, we offer a new foundation for robust and interpretable AI systems.

This research provides practical evidence of the bridge between quantum theory and artificial intelligence, opening new possibilities for guiding models in a world where the architecture learns the "physics" of the data itself. Rather than treating information as a collection of disjointed pixels or tokens, we treat it as a continuous wave-function governed by a learned potential landscape.

In the future, we will explore the application of these physical models to address the critical challenges of interpretability and hallucination in AI. We posit that:
\begin{itemize}
    \item \textbf{Dynamic Environment Robot Navigator.} The maze example in our experiment shows that the agent can dynamically adapt to real unseen environment thanks to its dynamic update mechanism at test time. This suggests the use case where environment can be changing over time as well, which directly apply to real world environment like in warehouse navigation or autonomous self-driving system, where obstacle are non-static.
      
    \item \textbf{General AI Agent.} In our paper, we have trained and tested a dual system AI Agent that solve the maze navigation. This shows the promising approach for Schr\"{o}dinger AI that can help solving a lot of difficult task for the current AI Agent such as hallucination, reliably generate output that align with facts and logics.

    \item \textbf{Personal AI.} As shown in our experiment, our model has the ability to adapt to new rule at test time, does not require retraining the whole model. This fact opens up the future work for personal AI where an AI agent with knowledge of the task or of the world can be taught at test time about new rule it needs to follow and new information about each specific user. 

    \item \textbf{Interpretability through Spectral Analysis}: By moving from "black-box" weights to a system of eigenvalues and wave-functions, we gain an analytical "fingerprint" for every decision the model makes. A model's confidence can be measured by the "spectral gap" between its axioms—providing a transparent, physically grounded metric for reliability.
\end{itemize}

\bibliographystyle{abbrv}
\bibliography{bib}

\appendix
\section{More example of maze solving agent in different mazes}

    \begin{figure}[b]
    \centering
    \includegraphics[width=\linewidth]{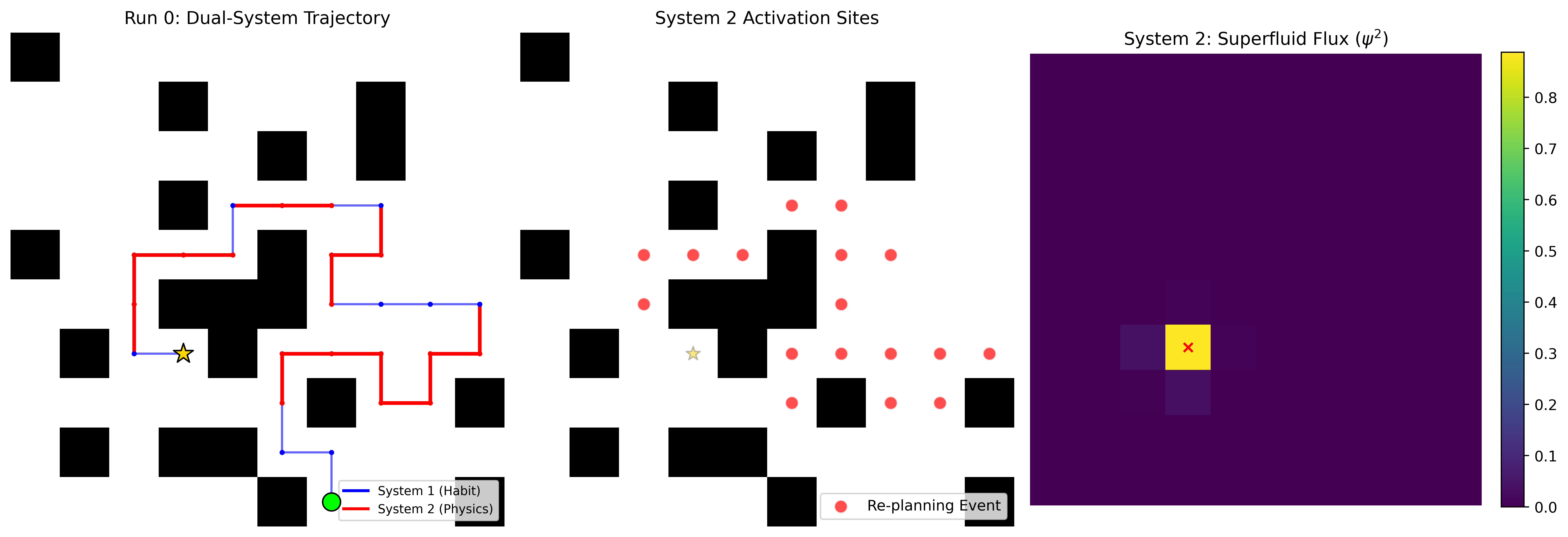}
    \end{figure}
    \begin{figure}[b]
    \centering
    \includegraphics[width=\linewidth]{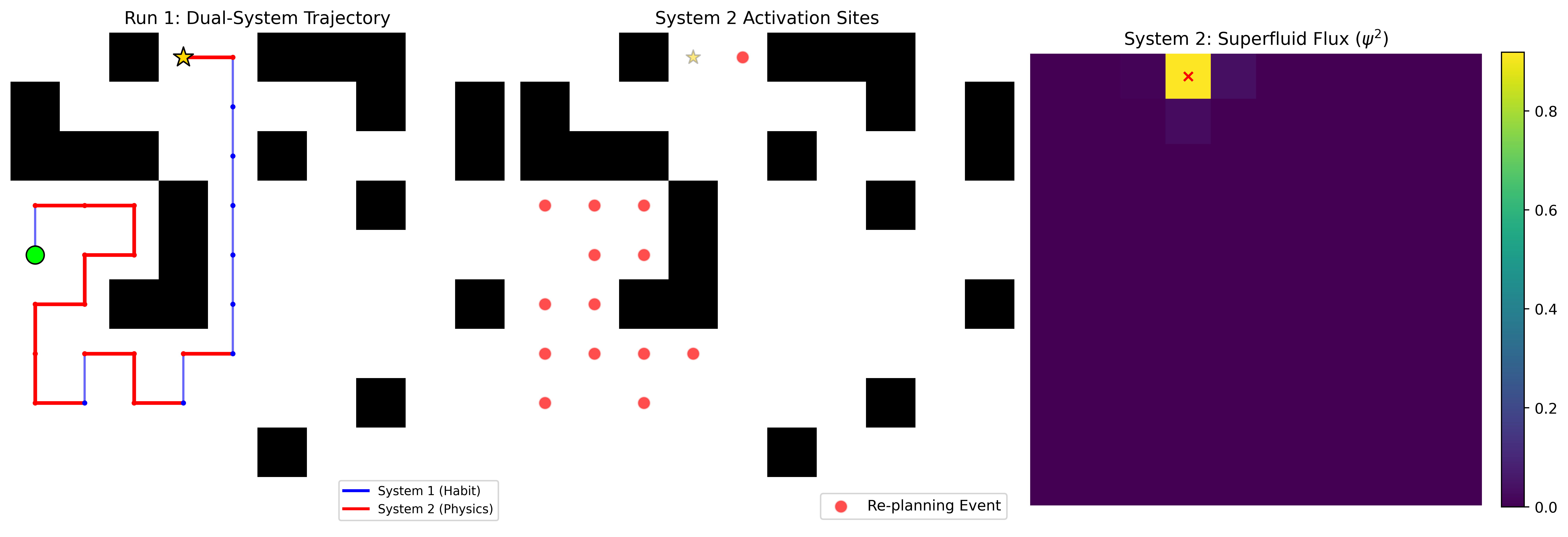}
    \end{figure}\begin{figure}[b]
    \centering
    \includegraphics[width=\linewidth]{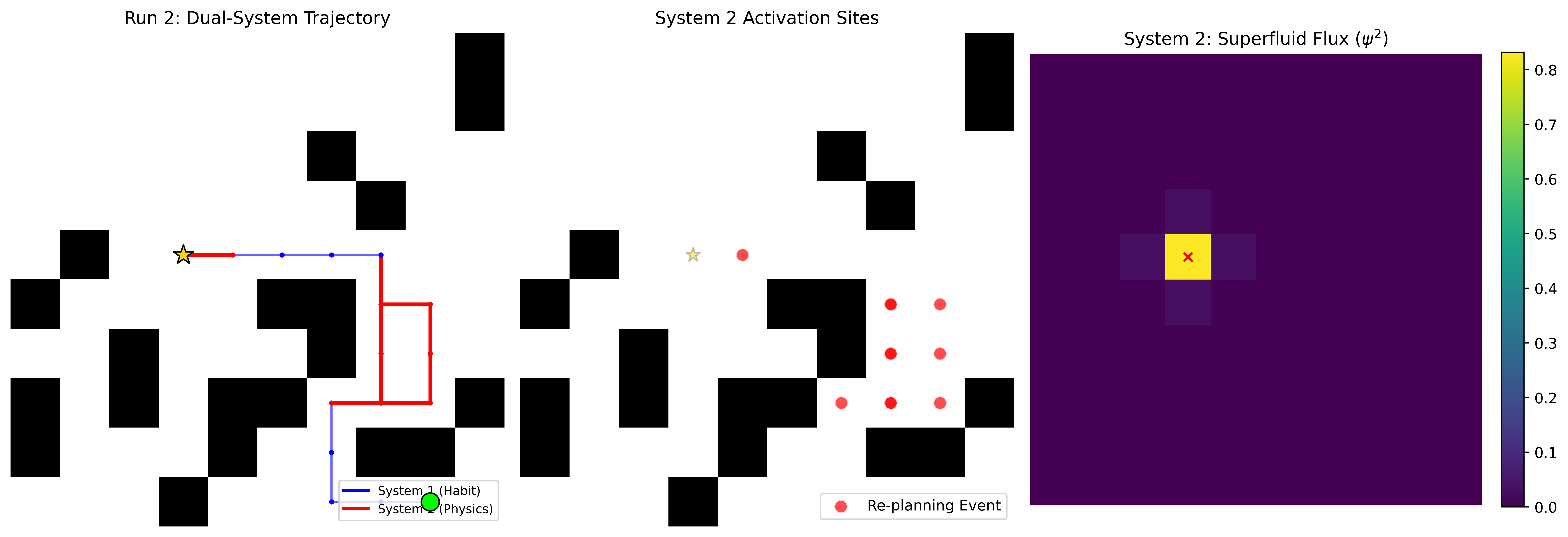}
    \end{figure}\begin{figure}[b]
    \centering
    \includegraphics[width=\linewidth]{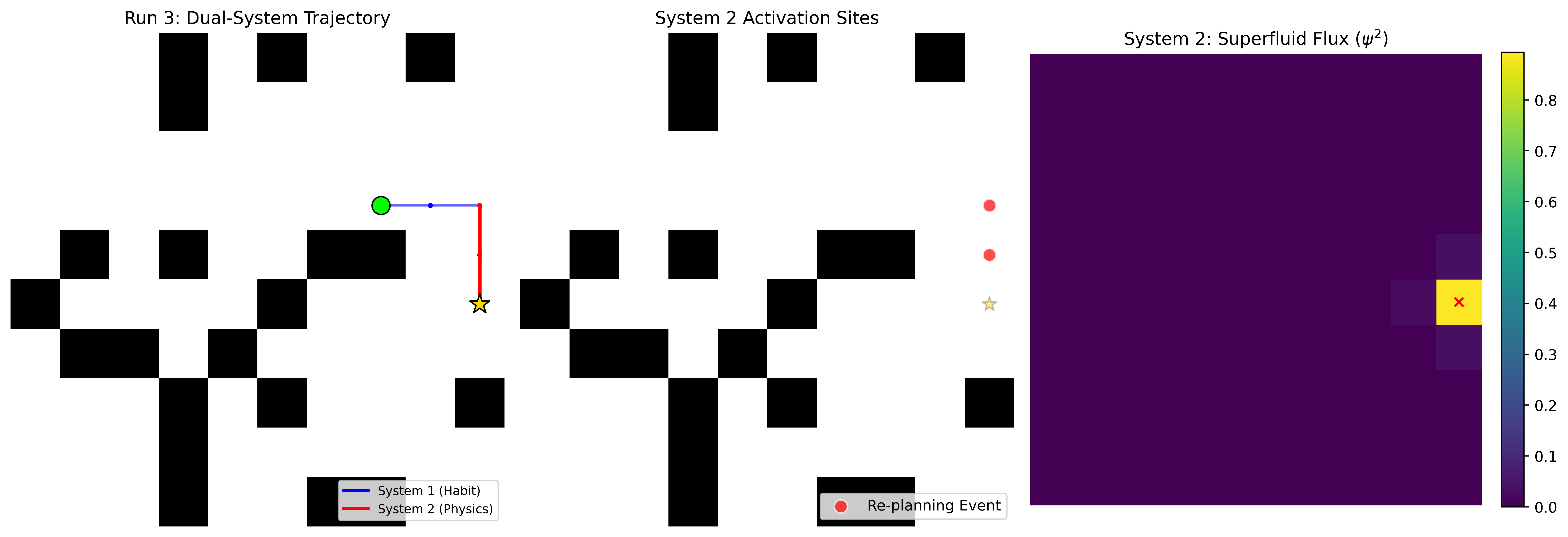}
    \end{figure}\begin{figure}[b]
    \centering
    \includegraphics[width=\linewidth]{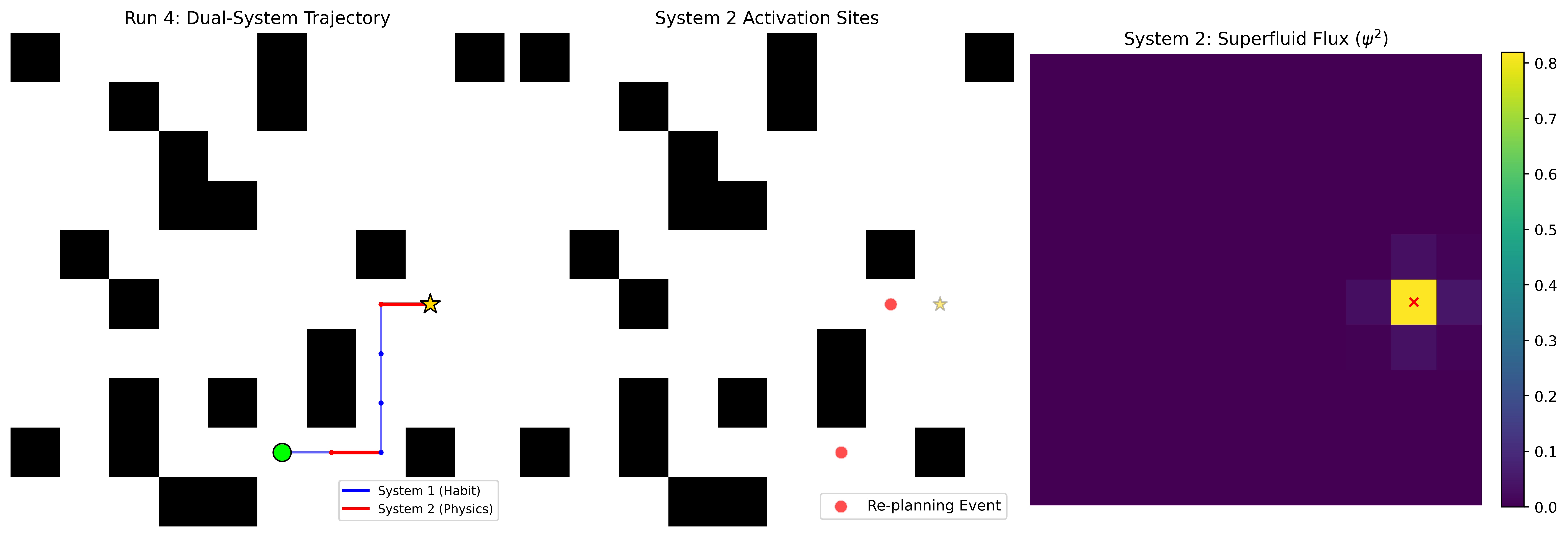}
    \end{figure}\begin{figure}[b]
    \centering
    \includegraphics[width=\linewidth]{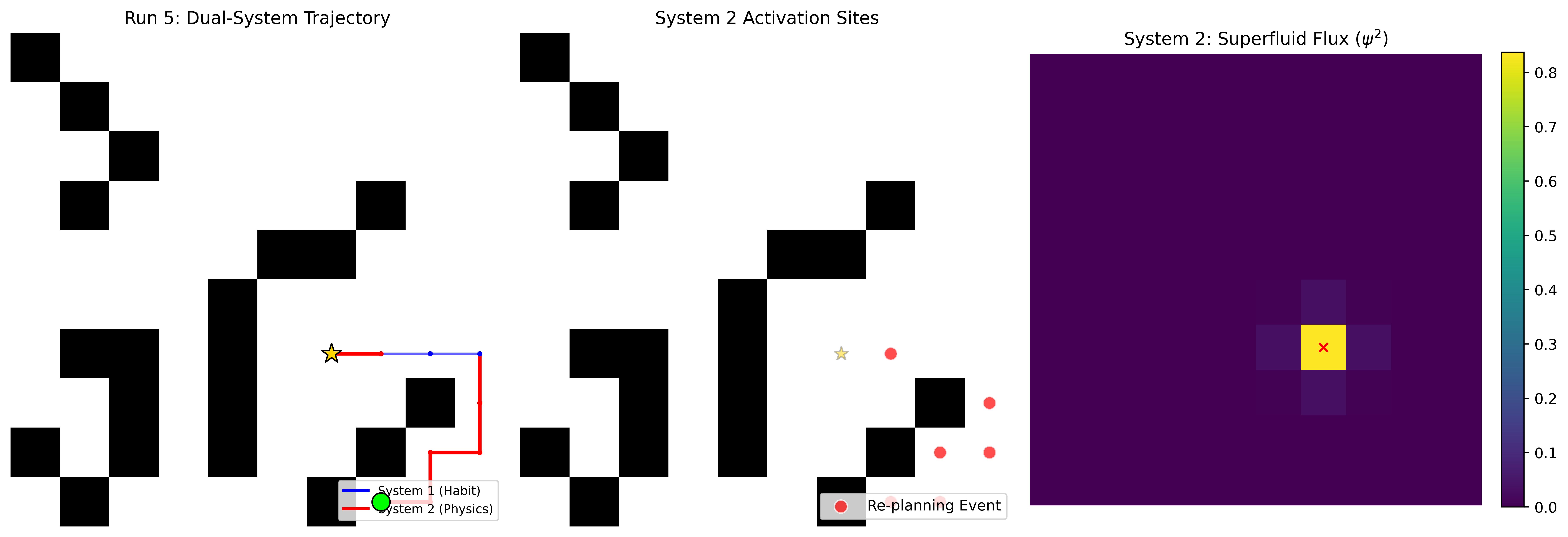}
    \end{figure}\begin{figure}[b]
    \centering
    \includegraphics[width=\linewidth]{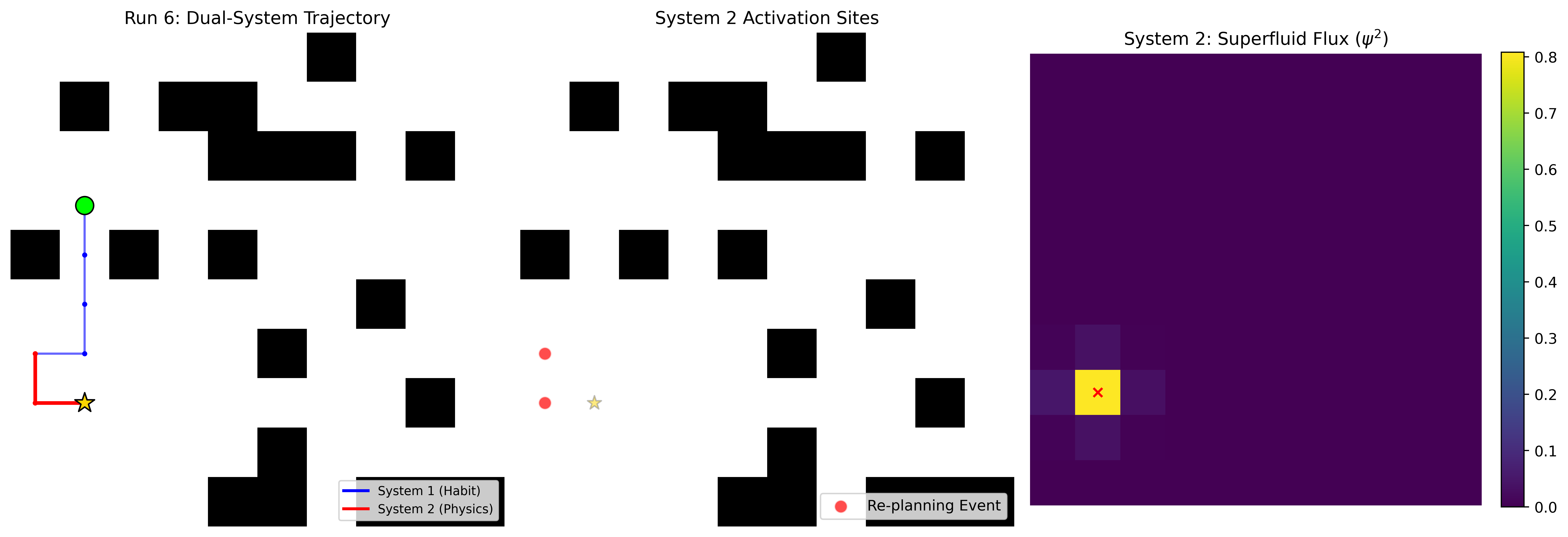}
    \end{figure}\begin{figure}[b]
    \centering
    \includegraphics[width=\linewidth]{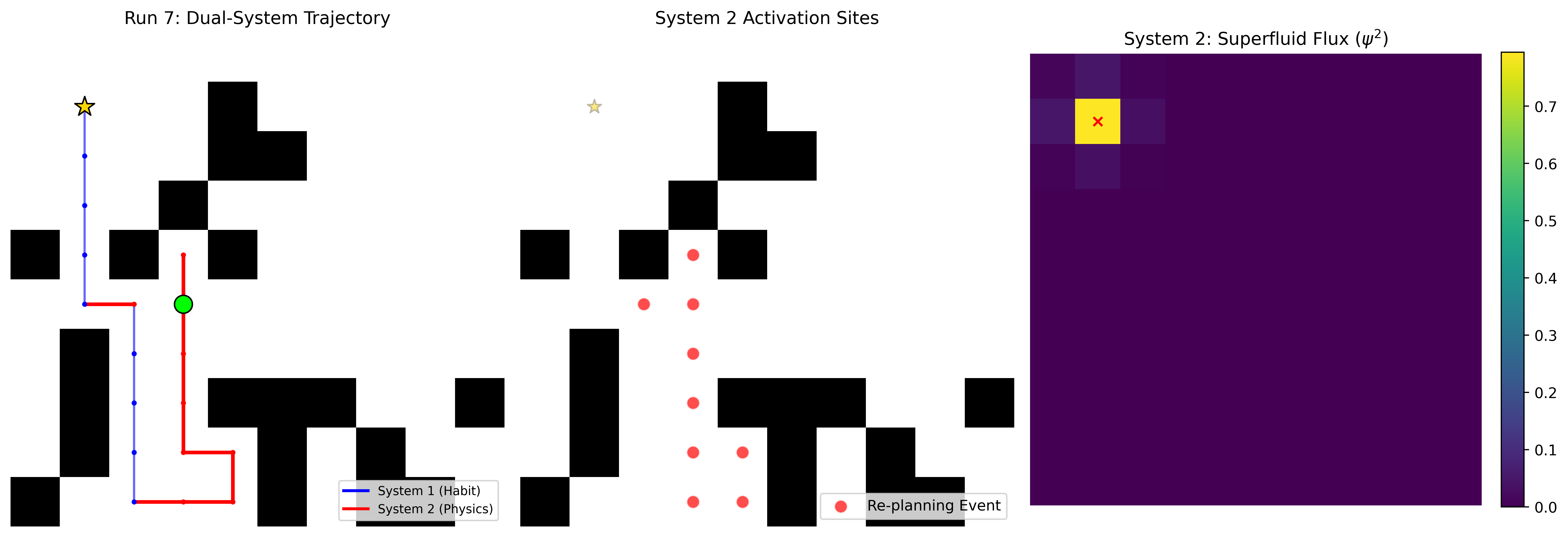}
    \end{figure}\begin{figure}[b]
    \centering
    \includegraphics[width=\linewidth]{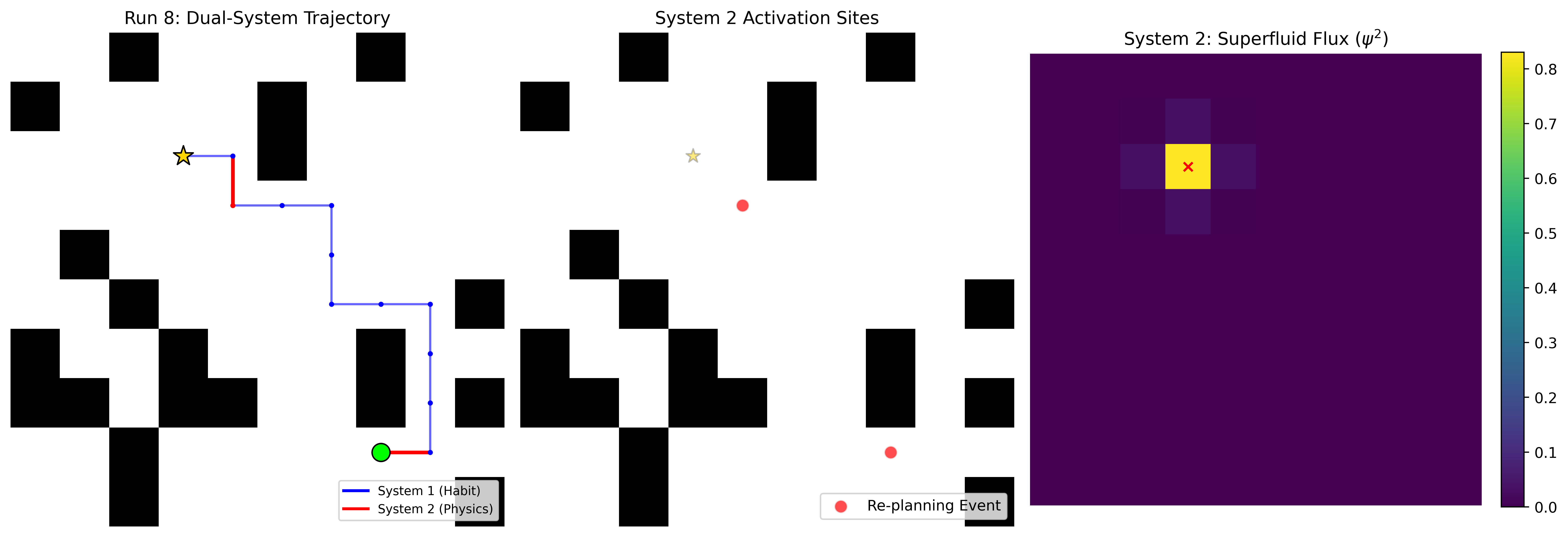}
    \end{figure}\begin{figure}[b]
    \centering
    \includegraphics[width=\linewidth]{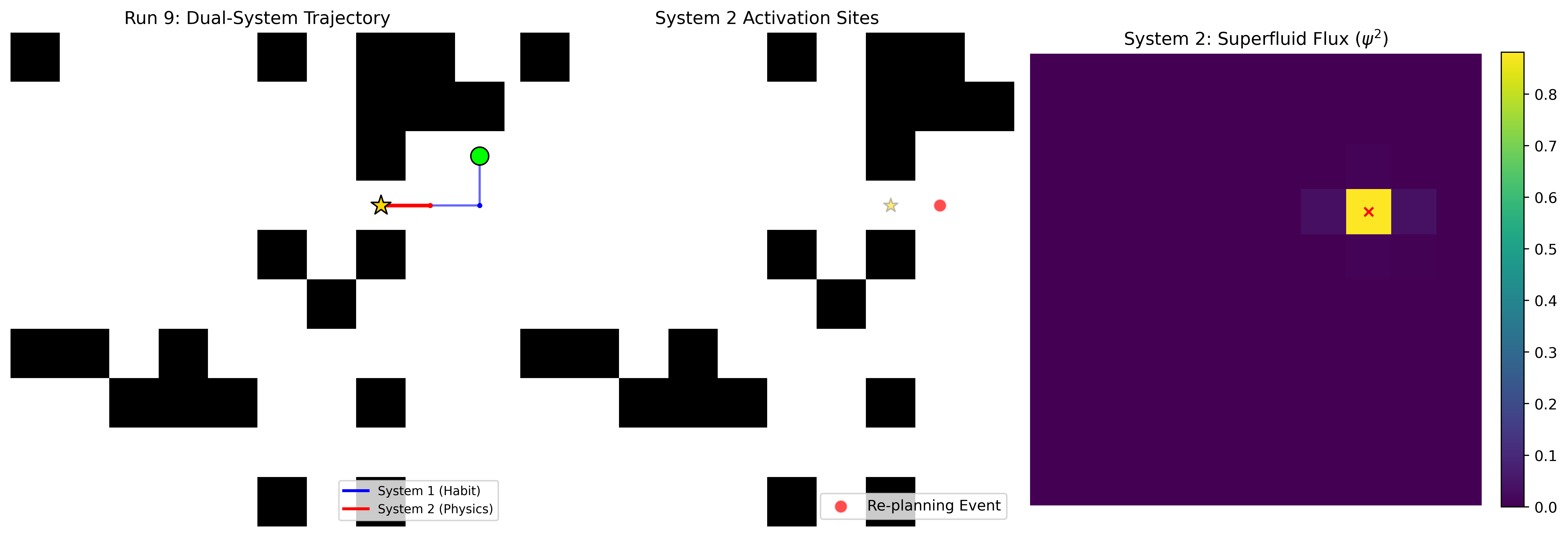}
    \end{figure}
    \begin{figure}[b]
    \centering
    \includegraphics[width=\linewidth]{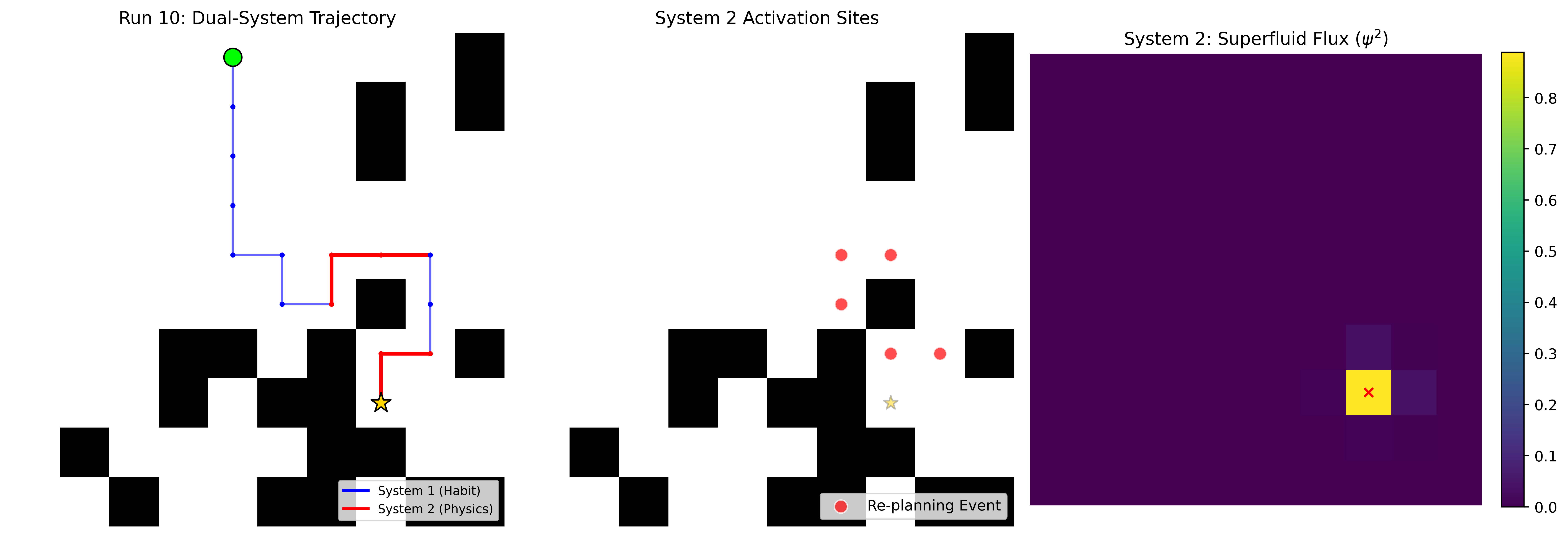}
    \end{figure}
    \begin{figure}[b]
    \centering
    \includegraphics[width=\linewidth]{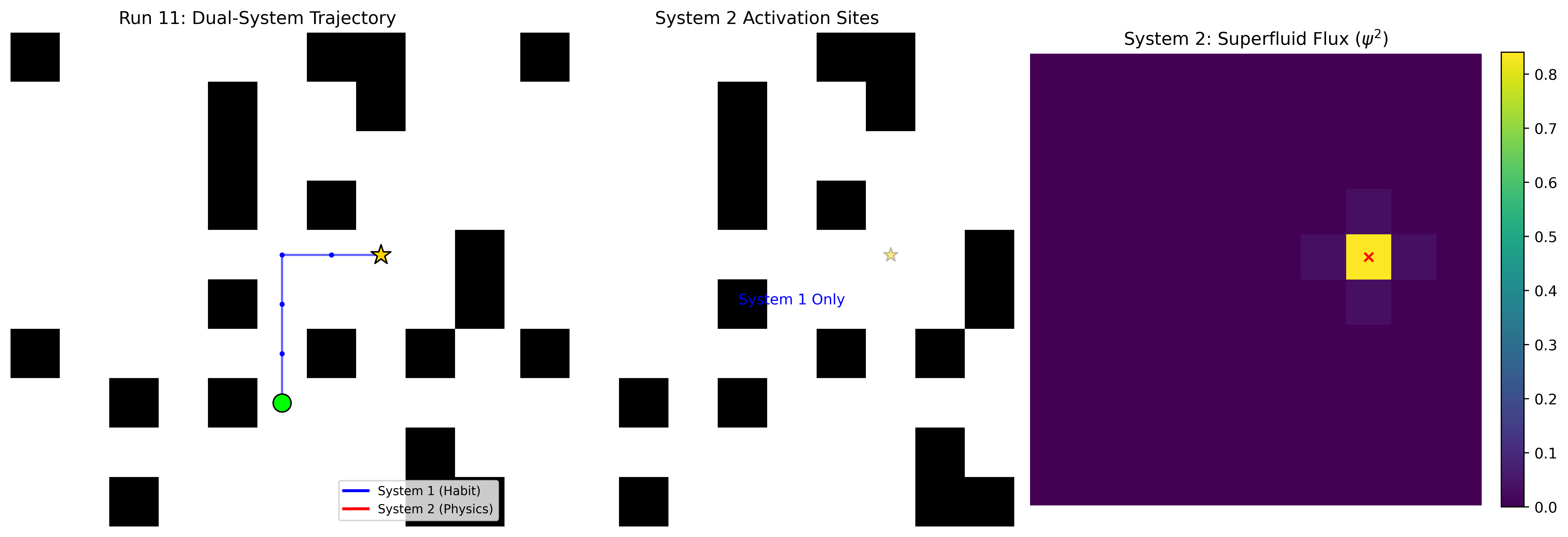}
    \end{figure}
    

\end{document}